\documentclass[letterpaper]{article} 
\usepackage{aaai24}  
\usepackage{times}  
\usepackage{helvet}  
\usepackage{courier}  
\usepackage[hyphens]{url}  
\usepackage{graphicx} 
\usepackage{natbib}  
\usepackage{caption} 
\usepackage{algorithm}
\usepackage{algorithmic}
\usepackage{booktabs}

\usepackage{newfloat}
\usepackage{listings}
\DeclareCaptionStyle{ruled}{labelfont=normalfont,labelsep=colon,strut=off} 
\floatstyle{ruled}
\newfloat{listing}{tb}{lst}{}
\floatname{listing}{Listing}
\nocopyright 

\title{Breaking Language Barriers: A Question-Answering Dataset for Hindi and Marathi}
\author{
    Maithili Sabane\equalcontrib \textsuperscript{\rm 1}, 
    Onkar Litake\equalcontrib \textsuperscript{\rm 2}, \\
    Aman Chadha\textsuperscript{\rm 3,4}\footnote{Work does not relate to position at Amazon.} 
}
\affiliations{
    \textsuperscript{\rm 1}University of Southern California, 
    \textsuperscript{\rm 2}University of California - San Diego, \\
    \textsuperscript{\rm 3}Stanford University,
    \textsuperscript{\rm 4}Amazon AI

}

\begin{document}

\maketitle

\begin{abstract}
The recent advances in deep-learning have led to the development of highly sophisticated systems with an unquenchable appetite for data. On the other hand, building good deep-learning models for low-resource languages remains a challenging task. This paper focuses on developing a Question Answering dataset for two such languages- Hindi and Marathi. Despite Hindi being the 3rd most spoken language worldwide, with 345 million speakers, and Marathi being the 11th most spoken language globally, with 83.2 million speakers, both languages face limited resources for building efficient Question Answering systems. To tackle the challenge of data scarcity, we have developed a novel approach for translating the SQuAD 2.0 dataset into Hindi and Marathi. We release the largest Question-Answering dataset available for these languages, with each dataset containing 28,000 samples. We evaluate the dataset on various architectures and release the best-performing models for both Hindi and Marathi, which will facilitate further research in these languages. Leveraging similarity tools, our method holds the potential to create datasets in diverse languages, thereby enhancing the understanding of natural language across varied linguistic contexts. Our fine-tuned models, code, and dataset will be made publicly available.
\end{abstract}

\section{Introduction}

Question Answering is one of the predominant tasks of Natural Language Processing that is concerned with developing systems that can automatically generate or retrieve an answer for the given questions from the text document. Unlike search systems that generate a series of related documents, QA models generate an answer from the knowledge base. These systems have a wide range of applications in information extraction and have gained major importance in the field of artificial intelligence due to the growth in the technical domain. 

These systems differ in the way answers are generated, such as: (i) extractive QA and (ii) generative QA. Extractive QA models generate answers directly from the provided knowledge base, and they often utilize various Transformer-based models such as BERT. In this approach, the answer is selected or copied from the original text, serving as a word marker for the extracted answer. On the other hand, generative QA models produce free text answers based on contextual information, leveraging text generation models. In this case, the answer is generated from scratch and is not constrained to any specific portion of the original text. Instead, it is an autonomously created response, distinguishing itself from the extractive approach. 

In addition to the differentiation between extractive and generative QA models, Question Answering systems can also be classified based on their domain of operation. This classification involves distinguishing between open-domain and closed-domain QA systems. Open-domain QA systems retrieve the answer from the context of the given document or a broader knowledge base. In other words, they are not limited to specific predefined answers and can provide responses that are not explicitly present in the original text. These systems are more flexible and capable of handling a wide range of questions, even those they have not encountered during training. On the other hand, closed-domain QA systems are designed to generate answers solely based on the knowledge contained within their training data. They are limited to providing responses within a predefined set of possible answers. Consequently, closed-domain QA systems may be more suitable for specific use cases with well-defined and structured questions and answers. In summary, open-domain QA systems have a broader scope and can generate responses beyond their training data, while closed-domain QA systems are more restricted in their answers and are tailored to specific domains with predefined sets of responses.

QA models are predominantly built and trained for the English language and further optimized with a large English corpus (e.g., SQuAD \cite{rajpurkar2016squad}). Such datasets are not available for Indian languages, and hence these are termed low-resource languages. The scarcity of sufficient corpora hinders the progress of NLP models for these low-resource languages.

India has a population of 1.4 billion and is home to 122 languages and 270 mother tongues. Indian languages fall in the minority as far as Natural Language Processing models are concerned. One of the Indian languages -- Hindi -- is spoken by 577.8 million people worldwide. Likewise, Marathi, which is another Indian language, is ranked 11th on the list of most spoken languages.  Despite the popularity of these languages, NLP systems in these languages remain unexplored.

While LLMs have made remarkable progress in natural language understanding, we'd like to highlight that using an extracted QA dataset in our research serves a specific purpose. These datasets provide a structured framework for benchmarking and evaluating various aspects of language models, not just their raw performance. Additionally, they remain relevant for research into data scarcity, model interpretability, fairness, and domain-specific applications. Our work aims to contribute insights into the limitations and strengths of traditional QA datasets and LLMs, fostering a comprehensive understanding of the field. We believe our research remains valuable even in the context of advanced LLMs.

Our works's novelty is grounded in creating a substantial QA dataset for Hindi and Marathi languages. These languages collectively have a vast user base of 714 million speakers. This contribution addresses a critical gap in the availability of resources for these languages. Even though SQuAD dataset has been present for a long period of time, still no work has been carried out for these low-resource languages. There are no Question-answering datasets and models for Marathi and Hindi. The SQuAD dataset set a precedent for evaluating language models and has led to numerous spin-off works that have expanded our understanding of reasoning abilities. In a similar way, our work aims to replicate the success of SQuAD by providing a foundational QA dataset for Hindi and Marathi. The impact of our work extends beyond the dataset itself. It paves the way for advancements in natural language understanding, reasoning, and machine learning applications for Hindi and Marathi-speaking communities.

To the best of the authors' knowledge, this dataset is the largest for Hindi and Marathi QA. Our approach tries to resolve the difficulty in accurately determining the index of the answer within the context. Our approach is explained in the Dataset Creation section.

Our contributions can be summarized as follows:
\begin{itemize}
 \item We present the largest Question-Answering dataset released for both Marathi and Hindi languages.
 \item We release the best-performing model for Question Answering based on our comprehensive evaluation and experimentation.
 \end{itemize}

    

\section{Related Work}

Question-Answering models \cite{allam2012question} are artificial intelligence or deep learning algorithms designed to respond to queries within specific contexts and, at times, even without any context. These models must possess a profound understanding of the language's structure and semantics, enabling them to locate the answer phrase within the given context accurately. However, to achieve such proficiency, these models require extensive training on large datasets. Manual creation of such datasets poses a significant challenge due to the time-consuming annotation process involved.

To address the scarcity of datasets for new languages, methods have been explored to create datasets based on existing ones. Translating the context and answers within the dataset is a relatively straightforward task. However, finding the exact index of the translated answer within the context becomes significantly more difficult. This challenge arises from the black-box nature of Language Models (LMs), where the translation of the answer, with or without surrounding context, can differ, making it challenging to directly determine the index at which the answer lies within the context.

The Dataset Creation section provides a more detailed explanation of the complexities encountered during this process. Consequently, constructing a model capable of delivering accurate answers becomes a formidable task due to the scarcity of datasets and the associated challenges that need to be addressed.

The availability of large-scale datasets, such as SQuAD \cite{squad}, Natural Questions (NQ) \cite{nq}, CoQA \cite{coqa}, and others, play a pivotal role in enhancing the model training process, especially for deep learning techniques. Deep Learning models have an insatiable data appetite; they require vast amounts of annotated data for effective training and performance improvement.

These large datasets offer a rich variety of real-world examples, covering diverse contexts and language patterns, allowing deep-learning models to learn from a wide range of scenarios. Consequently, the models can better grasp the complexities and nuances of natural language and, in turn, produce more accurate and robust results.

By leveraging large datasets, deep learning algorithms can extract intricate patterns, correlations, and semantic relationships, enabling them to provide more meaningful responses and perform better in various natural language processing tasks, including Question Answering. Therefore, the significance of large datasets cannot be understated, as they are instrumental in feeding the data hunger of deep learning techniques and substantially elevating their capabilities to comprehend and process natural language.

Most of these datasets are created for high-resource languages like English, German, etc., which in turn lead to the development of better quality models in those specific languages. Due to the lack of availability of good quality data in the low-resource languages, research and development is hampered. The amount of work done on low-resource Indian languages like Hindi and Marathi is inadequate in comparison to English and German.

\cite{gupta} in their work experimented with multilingual models on the task of Machine Comprehension, which is a sub-task of Question-Answering (QA) for Hindi and English. They experimented with mBERT on monolingual, zero-shot, and cross-lingual fine-tuning setups. \cite{kumar} fine-tuned an mBERT-based QA model using augmented data by translating and transliterating QA samples of the target language into other languages, which in turn helped to boost the performance of the models. 
\begin{figure*}[ht!]
  
  \frame{\includegraphics [scale= 0.65] {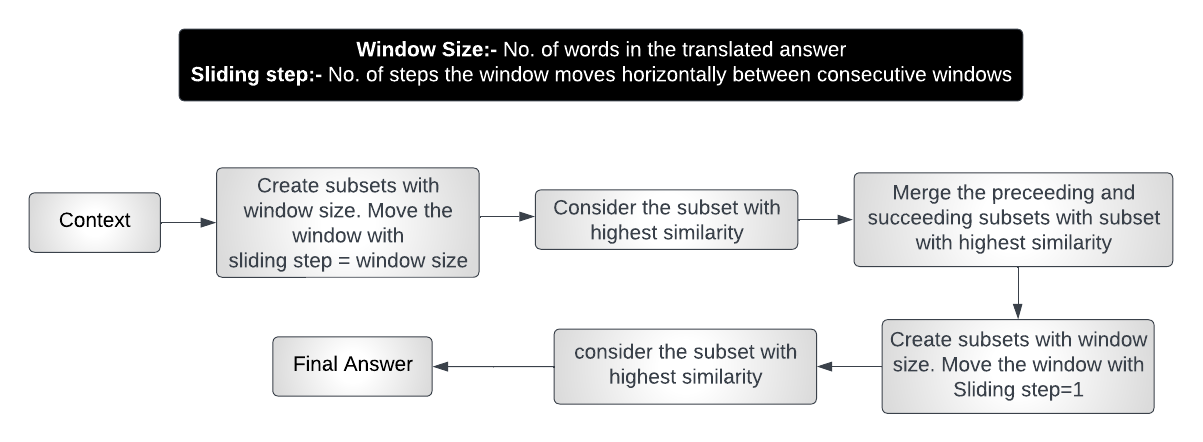}}
  \centering\caption{Sliding window technique for answer extraction.}
  \label{fig:1}
\end{figure*}
\cite{lewis} in order to tackle the difficulty of availability of datasets in languages other than English presented MLQA, a multi-way aligned extractive QA evaluation benchmark for excelling research in this area. It consisted of 12,000 QA instances in English and 5,0000 QA instances in 7 different languages, which included Hindi. \cite{gupta-deepak} curated a dataset containing 500 articles from 6 different domains. 250 of these articles consisted of English and Hindi documents, respectively. A total of 5,495 QA pairs were created. Difficulties in answering questions in multiple domains and languages were evaluated and necessary resources for developing a baseline model and bench-marking were assessed. 

\cite{artetxe} released the XQuAD dataset, which consisted of 240 paragraphs and 1190 question-answer pairs from SQuAD v1.1, which was translated by professional translators into ten different languages. The primary motivation behind this was to put forth a more comprehensive cross-lingual benchmark. \cite{anuranjana} released the first curated dataset, which was divided according to the level of difficulty of different grades for primary education. It consists of a total of 24 passages and 127 questions. As a part of the competition `Challenge in AI for India', Google Research released a dataset named `chaii-1' \cite{chaii}, which consists of QA pairs in Hindi and Tamil without the use of translation.

Although the scarcity of datasets for Hindi is conspicuous, some datasets do exist. MLQA (MultiLingual Question Answering \cite{lewis2019mlqa} is a benchmark dataset consisting of 5K extractive QA instances in seven languages, Hindi being one of them. The MMQA \cite{GUPTA18.826} dataset has around 6000 English-Hindi Question Answering pairs. HindiRC \cite{anuranjana2019hindirc} is a dataset for Reading Comprehension in Hindi containing 127 questions and 24 passages.

For the Marathi language, however, there exists no such public datasets to date. \cite{Govilkar} proposed a QA system for the Marathi language by making use of ontology. Domain-specific knowledge regarding semantic relationships and limitations in the relevant domains is expressed using ontologies. \cite{shelke} is working on curating datasets from textbooks for the primary grade classes. The dataset has a total of 901 questions. \cite{phade} proposed a system QA system for Marathi using transfer learning which utilizes the word embedding from a trained multilingual BERT model on a small custom dataset similar to SQuAD. \cite{kamble} developed a system for English-Marathi QA by manually scraping 1000 questions from Kaun Banega Crorepati (KBC) and translating them into Marathi. 

\section{Experimental Setup}
\subsection{Data Collection}
As the foundation of our research, we utilized the SQuAD dataset \cite{squad} to develop our Question-Answering system. It serves as a starting point for our work, and we build upon it to create a dataset specifically tailored for Marathi and Hindi languages. By leveraging the existing SQuAD dataset, we ensured a robust and standardized framework for the comparison and evaluation of our models' performance. This dataset comprises a collection of questions generated by crowd workers pertaining to a set of Wikipedia articles, with the goal of testing reading comprehension. Each question in the dataset requires an answer in the form of a specific text segment or span derived from the corresponding reading passage. Alternatively, some questions may not have an answer and are labeled as unanswerable.

\subsection{Dataset Creation}
Following is the structure of a single example of the SQuAD dataset:
    




We translate the given SQuAD dataset from English to Marathi and Hindi. We used IndicTrans \cite{tacl_a_00452} by AI4Bharat for Machine Translation. In order to develop a Question-Answering model, it is necessary to provide the character start index of the answer within the context. However, due to variations in sentence structures across different languages, the original dataset's indices cannot be directly utilized. To address this, we devised a novel method to determine the optimal indices accurately. Ensuring that the text in the answer aligns precisely with its appearance in the context poses a challenge due to the black-box nature of machine learning models, which can result in varied translations of the text.

During the compilation of the dataset, we address two primary challenges:
\begin{itemize}
    \item Determining the appropriate answer start index within the context.
    \item Replacing the translated answer with the exact corresponding answer as it appears in the context, ensuring precise alignment and maintaining the original meaning.
\end{itemize}

Figure \ref{fig:1} explains the process of extracting the final answer from the context. To identify the indices of the answer within the context, we employed a sliding window technique. This approach involved utilizing a window of length equal to the size of the answer text. By iterating through the context with this window size, we aimed to locate the subset of the context that exhibited the highest similarity to the answer text. This enabled us to pinpoint the precise indices corresponding to the answer within the context. We have used the MahaNLP \cite{joshi2022l3cube} library to calculate the similarity.
\begin{figure*}[ht!]
  
  \frame{\includegraphics [scale= 0.7] {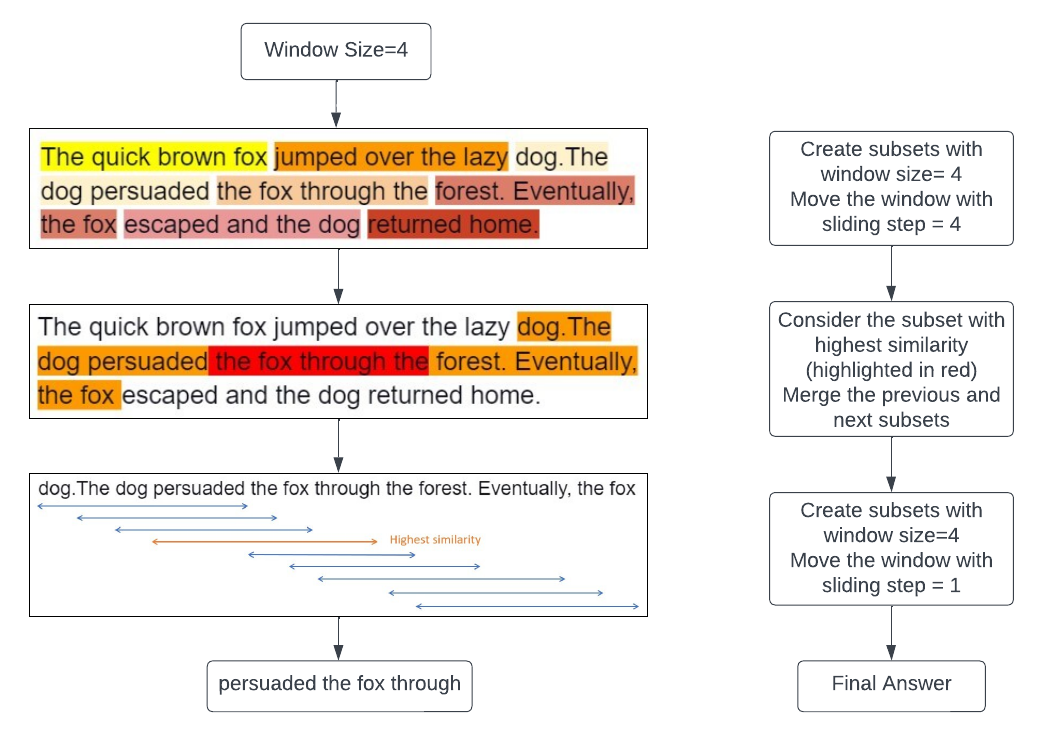}}
  \centering\caption{Illustration of the method using an English example.}
  \label{fig: 2}
\end{figure*}
To address the possibility of the final answer being spread across multiple subsets, we adopt a merging approach. After identifying the subset with the maximum similarity, we merge it with the subsets preceding and succeeding it. This consolidation allows us to consider the entire range where the answer might reside. To determine the final subset with maximum similarity, we traverse through the merged subsets with a window size similar to the previous approach, moving a distance of 1 at a time. Moreover, along with the character indices, we also calculate the token start and token end index of the answer, which can contribute to broader applications in model training and development. Additionally, once we obtain the index of the answer in the context, we replace the answer text with the subset exhibiting the highest similarity.  This methodology ensures comprehensive coverage of the answer while maintaining accuracy through the similarity-based selection process. 

The reason to use these specific models for translation and calculating similarity is that these are currently the only/best performing open source models according to our knowledge. The non-existent of multiple variant of these models again highlights the scarcity of resources for these 
low-resource languages.

Following is an example in English. The Translated Answer will originally be in target language(Hindi/Marathi), here it is kept in English for reader's convenience. 

\begin{itemize}

    \item 
    \textbf{Context:} ``The quick brown fox jumped over the lazy dog. The dog persuaded the fox through the forest. Eventually, the fox escaped and the dog returned home."

     \item 
    \textbf{Real Answer:} persuaded the fox through

    \item 
    \textbf{Translated Answer:} chased the fox through

   \item 
    \textbf{Window size:} length(Translated Answer) = 4
\end{itemize}





    

Figure \ref{fig: 2} provides an illustration of the sliding window technique for locating the final answer in the context.

The decision to employ a stride of window size was made with computational efficiency in mind. The need for such a process arises because the answer is paraphrased during translation and does not always lie in the context as it is. While the average context length is 78 words, it's worth noting that translation, especially for multiple examples, which totals around 28,000, can impose a significant computational load. Our approach aims to strike a balance between computational efficiency and adequate coverage of the data. 

We believe this approach efficiently assists us in locating the answer. We'd like to emphasize that even with a smaller step size from the start, we expect to obtain the same answer as with our mentioned approach.

\subsection{Dataset Statistics}

The dataset used for research was curated to encompass a wide array of textual data across different domains. Due to computational limitations, we translated 28,000 samples from SQuAD 2.0 using a machine translation model for Indic Languages. We use 21000 samples for training and 4200 samples each for testing and validation, respectively. 
Each sample is divided into title, context, question, answer, and answer start. On average, the context segments contain 78 words, while the question and answers have average lengths of 7 and 2 words, respectively. We observe similar trends in terms of context, question and answer lengths for test and validation sets. 

Moreover, an analysis of NER tags reveals additional insights into the dataset. Table \ref{result-ner} shows to count of each tag present in the answers for each of the train test and validation datasets. We use the MahaNLP library, trained on the L3Cube MahaNER \citep{litake-etal-2022-l3cube} dataset, to predict the NER tags.

\begin{table}
\begin{center}
\small
\begin{tabular}{lccc}
\toprule \textbf{Tag} & \textbf{Train} & \textbf{Test}  & \textbf{Valid}  \\
\midrule

     Date & 2113 & 464 & 413 \\
    
    Measure &  4495 & 927 & 974\\
    
    Organization & 1412 & 321 & 271  \\

    Location & 2069 & 459 & 466\\
    
    Person & 3415 & 606 & 697 \\
    
    Designation &  136 & 18 & 29\\
    Time & 59 & 8 & 14 \\
    \bottomrule
    
\end{tabular}
\caption{\label{result-ner}Count of NER tags present in the answer text.}
\end{center}
\end{table}

\begin{table*}
\begin{center}
\small
\begin{tabular}{lccccc}
\toprule \textbf{Model} & \textbf{EM(\%)} & \textbf{Rouge-2} & \textbf{Rouge-L} & \textbf{BLEU (Unigram)(\%)} &\textbf{BLEU (Bigrams)(\%)} \\
\midrule

     mBERT & 36.3 & 0.31 & 0.58 & 49.67 & 30.54 \\
    XLM-RoBERTa &  43.16 & 0.33 & 0.63 & 55.98 & 33.87\\
    Distil-BERT & 31.14 & 0.27 & 0.52 & 41.41 & 25.26 \\
    Hindi BERT & \textbf{47.84} & \textbf{0.36} & \textbf{0.66} & \textbf{61.47} & \textbf{38.51}\\
    Hindi RoBERTa & 43.33 & 0.34 & 0.63 & 56.28 & 34.44 \\
    \bottomrule
    
\end{tabular}
\caption{\label{result-tab-hindi}Exact Match (EM), Rouge-N, Rouge-L, BLEU (Unigram), BLEU (Bigrams) for various models on Hindi dataset.}
\end{center}
\end{table*}

\begin{table*}
\begin{center}
\small
\begin{tabular}{lccccc}
\toprule \textbf{Model} & \textbf{EM(\%)} & \textbf{Rouge-2} & \textbf{Rouge-L} & \textbf{BLEU (Unigram)(\%)} &\textbf{BLEU (Bigrams)(\%)} \\
\midrule

     mBERT & 28.78 & 0.32 & 0.54 & 43.34 & 30.34 \\
    XLM-RoBERTa &  37.92 & 0.34 & 0.57 & 47.65 & 31.42\\
    Distil-BERT & 25.11 & 0.27 & 0.46 & 35.59 & 24.00  \\
    MahaBERT & \textbf{42.97} & \textbf{0.38} & \textbf{0.62} & \textbf{57.42} & \textbf{39.70}\\
    MahaRoBERTa & 38.52 & 0.35 & 58 & 52.03 & 35.64 \\
    \bottomrule
    
\end{tabular}
\caption{\label{result-tab-marathi} Exact Match (EM), Rouge-N, Rouge-L, BLEU (Unigram), BLEU (Bigrams) for various models on Marathi dataset.}
\end{center}
\end{table*}

\section{Experimentation}

\subsection{Model Selection}

Transformer-based \cite{vaswani2023attention} models have dominated the field of natural language processing due to their remarkable ability to capture complex linguistic relationships.
This architecture's capacity for bidirectional context understanding allows it to grasp contextual nuances and dependencies within the text, making it highly effective for tasks like language translation, sentiment analysis, and question answering.  Moreover, as they are pre-trained on large corpora and subsequently fine-tuned on task-specific data, transformers contribute to their widespread adoption and consistent achievement of state-of-the-art results across various benchmarks.

To establish a baseline we have used several well-known LMs, including monolingual and multilingual. Following are the general LMs used for training on our dataset:

\textbf{mBert: }mBERT \citep{mbert} stands for Multilingual BERT, is a variant of the BERT model pre-trained on 104 languages that makes it adaptable to various multilingual tasks, including Question Answering. Through its tokenization, segment embedding, and positional encoding techniques, mBERT can precisely capture the intricate structures of questions and passages.  

\textbf{XLM-RoBERTa: }XLM-RoBERTa \citep{xlm} is an evolution of the original BERT model and makes several modifications to the training procedure, leading to improved performance. Some of the changes include larger batch size, longer training time, and larger input. It is an extension of Roberta \citep{liu2019roberta} that is designed to handle multilingual and cross-lingual tasks. It extends the RoBERTa architecture to handle cross-lingual tasks by learning shared representations of words in different languages.

\textbf{DistilBERT: }DistilBERT \cite{Sanh2019DistilBERTAD} is a distilled version of the BERT base multilingual model. DistilBERT reduces the usage of computational resources by using a distillation technique that transfers the knowledge from the larger BERT model to a smaller one. It achieves faster inference times and requires less memory, making it well-suited for deployment in resource-constrained environments. 

The Marathi language-specific models utilized for training on our dataset were introduced by \cite{joshi-2022-l3cube} in their recent research. These models consist of \textbf{MahaBERT}, a Marathi BERT model, which is a fine-tuned version of the multilingual BERT-base-cased model. \textbf{MahaRoBERTa}, a Marathi RoBERTa model, which is based on the multilingual RoBERTa (\texttt{xlm-roberta-base}) model. All of these models are fine-tuned using L3Cube-MahaCorpus and other publicly available Marathi monolingual datasets.

Similarly, the Hindi language-specific models utilized for training on our dataset were introduced by \cite{joshi2022l3cubehind}. These models consist of \textbf{HindBERT}, a Hindi BERT model, which is a fine-tuned version of the multilingual BERT-base-cased model. \textbf{HindRoBERTa}, a Hindi RoBERTa model, which is based on the multilingual RoBERTa (xlm-roberta-base) model. All of these models are fine-tuned on publicly available Hindi monolingual datasets. 

The decision to focus on BERT-based models aligns with the primary motivation of our paper, which was to create a comprehensive dataset for Question and Answering (QnA) tasks. We used BERT-based models for evaluation primarily to ensure the correctness and effectiveness of our dataset. These models have established themselves as a benchmark for natural language understanding, and by using them, we aimed to guarantee the reliability and generalizability of our dataset.

To keep up with the ongoing trend on usage of LLMs, we were not able to employ them as according to our knowledge currently there are no open-source LLM for low-resource languages such as Marathi and Hindi. Our extensive survey of the existing multilingual LLMs has
shown that their performance is suboptimal for various tasks
in these languages.

\subsection{Experimental Settings}

We fine-tuned the aforementioned models on our dataset using PyTorch \cite{paszke2019pytorch}. The models were fine-tuned using a single Tesla T4 GPU, and 12 GB RAM. A batch size of 2 was employed, and optimization was performed using the AdamW optimizer \cite{kingma2017adam} with a learning rate of 1e-5. The training process spanned five epochs for all the models, with gradient accumulation taking place over two batches. These hyperparameters collectively shaped the training dynamics and contributed to the outcomes of the experiments.

\section{Results}

In this study, we have experimented with various monolingual and multi-lingual LMs on our dataset for Hindi and Marathi. We have used the following evaluation metrics: 

\textbf{Exact Match:} It measures the percentage of instances where the predicted output exactly matches the reference output.

\textbf{Rouge-2: }ROUGE \cite{rouge} (Recall-Oriented Understudy for Gisting Evaluation) is used for evaluating the quality of machine-generated text. Rouge-N is one specific metric within the Rouge family, where ``N" refers to the size of the n-grams being compared. Here we have calculated the Rouge-2 score, which indicates that bi-grams are compared. The reason for choosing a bi-gram is stated at the end of the section.

\textbf{Rouge-L: }ROUGE-L measures the longest common subsequence (LCS) between the predicted text and the reference text. The LCS represents the longest sequence of words that appears in both the candidate and reference text. Rouge score are represented between scale of 0 to 1.

\textbf{BLEU: }BLEU \cite{bleu} (Bilingual Evaluation Understudy) measures the similarity between the machine-generated output and one or more human reference outputs. BLEU calculates precision scores based on overlapping n-grams. Here we have selected $n=2$ to calculate overlapping unigram and bi-grams. The reason for choosing n to be two is stated at the end of the section.

As mentioned in the `Dataset Statistics' section, the average number of words for `answer' (predicted text) is 2; hence we decided to keep $n=2$ for Rouge and BLEU scores. The results have been reported in Table \ref{result-tab-hindi} and Table \ref{result-tab-marathi}, respectively. Hindi BERT outperforms the other models on the Hindi dataset; similarly `MahaBERT' model performs best on the Marathi dataset.

The reason for Indic models outperforming general models is that these language-specific models focus solely on one language, such as Marathi or Hindi, in contrast to general models designed to work with many languages. This dedicated training approach helps them understand the unique aspects of that language – things like how words are used, sentence structures, and specific context details that are part of the language's nature. It makes the model knowledgeable about specific topics, phrases, and everyday words used for that specific languages. In short, the success of models like MahaBERT and Hindi BERT comes from their focused training on a particular language.

\section{Conclusion}

We introduce the biggest dataset for Question-Answering in Marathi and Hindi languages consisting of 28,000 samples. This is the first step taken to reduce the missing literature in existing research work. With the availability of models for Machine Translation, and Sentence Similarity, the same method can be used to create datasets for different Indic languages. Hindi BERT performs best for QA on the Hindi dataset; similarly, `MahaBERT' performs best for QA on the Marathi dataset. We plan to release these models along with our dataset. We want to encourage researchers to build better models for QA tasks. We think that the development of conversational AI for the Marathi language will greatly benefit from the use of our corpus.

\section{Reproducibility}
The pre-trained models, dataset, and code will be made publicly available. 


\end{document}